\documentclass[letterpaper, 10 pt, journal, twoside]{ieeetran} 

\IEEEoverridecommandlockouts                              

\usepackage{times}
\usepackage{multicol}
\usepackage[bookmarks=true]{hyperref}

\usepackage{caption}
\usepackage{times}

\usepackage{amssymb,amsfonts,amsmath,amscd}
\usepackage[pdftex]{graphicx}
\usepackage{fnpos}
\usepackage[bf]{subfigure}
\usepackage{verbatim}
\usepackage{fancyhdr}
\usepackage{pgfgantt}
\usepackage{mathrsfs}
\usepackage{nicefrac} 
\usepackage{xcolor}
\usepackage{acronym}
\usepackage[font=footnotesize]{caption}
\usepackage[noadjust]{cite}

\usepackage{tikz}
\tikzset{
	font=\footnotesize}

\newcommand{\Real}{\mathbb R}
\newcommand{\bbm}{\begin{bmatrix}}
\newcommand{\ebm}{\end{bmatrix}}
\newcommand{\mbf}{\mathbf}
\newcommand{\mbs}[1]{{\boldsymbol{#1}}}

\newcommand{\beq}{\begin{equation}}
\newcommand{\eeq}{\end{equation}}
\newcommand{\bdis}{\begin{displaymath}}
\newcommand{\edis}{\end{displaymath}}
\newcommand{\beqn}[1]{\begin{subequations}\label{eq:#1}\begin{eqnarray}}
\newcommand{\eeqn}{\end{eqnarray}\end{subequations}}
\newcommand{\Tsmall}{\mbf{T}}

\newcommand{\bs}{\boldsymbol}
\newcommand{\jac}{\bs{\mathcal{J}}}

\newcommand{\varpii}{\bs{\varpi}}

\acrodef{ESGVI}{Exactly Sparse Gaussian Variational Inference}
\acrodef{KL}{Kullback-Leibler}
\acrodef{IW}{Inverse-Wishart}
\acrodef{EM}{Expectation Maximization}
\acrodef{GEM}{Generalized Expectation Maximization}
\acrodef{IRLS}{iteratively reweighted least squares}
\acrodef{WNOA}{white-noise-on-acceleration}
\acrodef{MAP}{Maximum A Posteriori}
\acrodef{SLAM}{Simultaneous Localization and Mapping}
\acrodef{ELBO}{Evidence Lower Bound}

\title{Variational Inference with Parameter Learning Applied to Vehicle Trajectory Estimation
}

\author{Jeremy N. Wong$^{1}$, David J. Yoon$^{1}$, Angela P. Schoellig$^{1}$, and Timothy D. Barfoot$^{1}$%
	\thanks{Manuscript received: February 24, 2020; Revised May 20, 2020; Accepted June 4, 2020.}
	\thanks{This paper was recommended for publication by Editor Sven Behnke upon evaluation of the Associate Editor and Reviewers' comments. 
		This work was supported by in part by Applanix Corporation and in part by the Natural Sciences and Engineering Research Council of Canada.} 
	\thanks{$^{1}$The authors are with the University of Toronto Institute for Aerospace Studies, University of Toronto, and with the Vector Institute for Artificial Intelligence, Toronto, Canada.
		{\tt\footnotesize \{jeremy.wong, david.yoon\}@robotics.utias.utoronto.ca, \{angela.schoellig, tim.barfoot\}@utoronto.ca}}%
	\thanks{Digital Object Identifier (DOI): see top of this page.}
}

\begin{document}

\maketitle

\begin{abstract}


We present parameter learning in a Gaussian variational inference setting using only noisy measurements (i.e., no groundtruth).  This is demonstrated in the context of vehicle trajectory estimation, although the method we propose is general.  The paper extends the Exactly Sparse Gaussian Variational Inference (ESGVI) framework, which has previously been used for large-scale nonlinear batch state estimation. Our contribution is to additionally learn parameters of our system models (which may be difficult to choose in practice) within the ESGVI framework. In this paper, we learn the covariances for the motion and sensor models used within vehicle trajectory estimation.  Specifically, we learn the parameters of a white-noise-on-acceleration motion model and the parameters of an Inverse-Wishart prior over measurement covariances for our sensor model.  We demonstrate our technique using a 36~km dataset consisting of a car using lidar to localize against a high-definition map; we learn the parameters on a training section of the data and then show that we achieve high-quality state estimates on a test section, even in the presence of outliers. Lastly, we show that our framework can be used to solve pose graph optimization even with many false loop closures.
\end{abstract}

\begin{IEEEkeywords}
	SLAM, Localization
\end{IEEEkeywords}

\section{Introduction}

\IEEEPARstart{P}{robabilistic} state estimation is a core component of mobile robot navigation. While the estimation machinery is reasonably mature, there are robot model parameters that are difficult to determine from first principles and vary with each new platform and sensor. Our vision is to develop a learning framework that allows the deployment of a robot with arbitrary sensors onboard, and have it learn the model parameters required for estimation (and planning/control) solely from the sensor data.  This can be viewed as a form of nonlinear system identification, although we will approach the problem using modern machine learning techniques.

\begin{figure}[t]
	\centering
	\vspace{1mm}
	\tikzset{every picture/.style={line width=0.75pt}} 
	\tikzset{every picture/.style={line width=0.75pt}} 

\begin{tikzpicture}[x=0.75pt,y=0.75pt,yscale=-1,xscale=1]

\draw   (70,140) .. controls (70,128.95) and (78.95,120) .. (90,120) .. controls (101.05,120) and (110,128.95) .. (110,140) .. controls (110,151.05) and (101.05,160) .. (90,160) .. controls (78.95,160) and (70,151.05) .. (70,140) -- cycle ;

\draw   (150,140) .. controls (150,128.95) and (158.95,120) .. (170,120) .. controls (181.05,120) and (190,128.95) .. (190,140) .. controls (190,151.05) and (181.05,160) .. (170,160) .. controls (158.95,160) and (150,151.05) .. (150,140) -- cycle ;

\draw    (110,140) -- (150,140) ;
\draw  [fill={rgb, 255:red, 28; green, 27; blue, 27 }  ,fill opacity=1 ] (132.75,140) .. controls (132.75,138.62) and (131.52,137.5) .. (130,137.5) .. controls (128.48,137.5) and (127.25,138.62) .. (127.25,140) .. controls (127.25,141.38) and (128.48,142.5) .. (130,142.5) .. controls (131.52,142.5) and (132.75,141.38) .. (132.75,140) -- cycle ;
\draw   (310,140) .. controls (310,128.95) and (318.95,120) .. (330,120) .. controls (341.05,120) and (350,128.95) .. (350,140) .. controls (350,151.05) and (341.05,160) .. (330,160) .. controls (318.95,160) and (310,151.05) .. (310,140) -- cycle ;

\draw  [fill={rgb, 255:red, 30; green, 29; blue, 29 }  ,fill opacity=1 ] (239.6,140.33) .. controls (239.6,140.04) and (239.84,139.8) .. (240.13,139.8) .. controls (240.41,139.8) and (240.65,140.04) .. (240.65,140.33) .. controls (240.65,140.61) and (240.41,140.85) .. (240.13,140.85) .. controls (239.84,140.85) and (239.6,140.61) .. (239.6,140.33) -- cycle ;
\draw  [fill={rgb, 255:red, 30; green, 29; blue, 29 }  ,fill opacity=1 ] (249.6,140.33) .. controls (249.6,140.04) and (249.84,139.8) .. (250.13,139.8) .. controls (250.41,139.8) and (250.65,140.04) .. (250.65,140.33) .. controls (250.65,140.61) and (250.41,140.85) .. (250.13,140.85) .. controls (249.84,140.85) and (249.6,140.61) .. (249.6,140.33) -- cycle ;
\draw  [fill={rgb, 255:red, 30; green, 29; blue, 29 }  ,fill opacity=1 ] (259.6,140.33) .. controls (259.6,140.04) and (259.84,139.8) .. (260.13,139.8) .. controls (260.41,139.8) and (260.65,140.04) .. (260.65,140.33) .. controls (260.65,140.61) and (260.41,140.85) .. (260.13,140.85) .. controls (259.84,140.85) and (259.6,140.61) .. (259.6,140.33) -- cycle ;

\draw    (270,140) -- (310,140) ;
\draw    (190,140) -- (230,140) ;
\draw   (70,220) .. controls (70,208.95) and (78.95,200) .. (90,200) .. controls (101.05,200) and (110,208.95) .. (110,220) .. controls (110,231.05) and (101.05,240) .. (90,240) .. controls (78.95,240) and (70,231.05) .. (70,220) -- cycle ;

\draw   (150,220) .. controls (150,208.95) and (158.95,200) .. (170,200) .. controls (181.05,200) and (190,208.95) .. (190,220) .. controls (190,231.05) and (181.05,240) .. (170,240) .. controls (158.95,240) and (150,231.05) .. (150,220) -- cycle ;

\draw   (310,220) .. controls (310,208.95) and (318.95,200) .. (330,200) .. controls (341.05,200) and (350,208.95) .. (350,220) .. controls (350,231.05) and (341.05,240) .. (330,240) .. controls (318.95,240) and (310,231.05) .. (310,220) -- cycle ;

\draw    (90,160) -- (90,200) ;
\draw  [fill={rgb, 255:red, 28; green, 27; blue, 27 }  ,fill opacity=1 ] (92.75,180) .. controls (92.75,178.62) and (91.52,177.5) .. (90,177.5) .. controls (88.48,177.5) and (87.25,178.62) .. (87.25,180) .. controls (87.25,181.38) and (88.48,182.5) .. (90,182.5) .. controls (91.52,182.5) and (92.75,181.38) .. (92.75,180) -- cycle ;

\draw    (170.75,159.67) -- (170.75,199.67) ;
\draw  [fill={rgb, 255:red, 28; green, 27; blue, 27 }  ,fill opacity=1 ] (173.5,179.67) .. controls (173.5,178.29) and (172.27,177.17) .. (170.75,177.17) .. controls (169.23,177.17) and (168,178.29) .. (168,179.67) .. controls (168,181.05) and (169.23,182.17) .. (170.75,182.17) .. controls (172.27,182.17) and (173.5,181.05) .. (173.5,179.67) -- cycle ;

\draw    (329.67,159.67) -- (329.67,199.67) ;
\draw  [fill={rgb, 255:red, 28; green, 27; blue, 27 }  ,fill opacity=1 ] (332.42,179.67) .. controls (332.42,178.29) and (331.19,177.17) .. (329.67,177.17) .. controls (328.15,177.17) and (326.92,178.29) .. (326.92,179.67) .. controls (326.92,181.05) and (328.15,182.17) .. (329.67,182.17) .. controls (331.19,182.17) and (332.42,181.05) .. (332.42,179.67) -- cycle ;

\draw    (90,240) -- (90,260) ;
\draw  [fill={rgb, 255:red, 28; green, 27; blue, 27 }  ,fill opacity=1 ] (92.83,260.5) .. controls (92.83,259.12) and (91.6,258) .. (90.08,258) .. controls (88.56,258) and (87.33,259.12) .. (87.33,260.5) .. controls (87.33,261.88) and (88.56,263) .. (90.08,263) .. controls (91.6,263) and (92.83,261.88) .. (92.83,260.5) -- cycle ;
\draw    (169.5,240.29) -- (169.5,260.29) ;
\draw  [fill={rgb, 255:red, 28; green, 27; blue, 27 }  ,fill opacity=1 ] (172.33,260.79) .. controls (172.33,259.41) and (171.1,258.29) .. (169.58,258.29) .. controls (168.06,258.29) and (166.83,259.41) .. (166.83,260.79) .. controls (166.83,262.17) and (168.06,263.29) .. (169.58,263.29) .. controls (171.1,263.29) and (172.33,262.17) .. (172.33,260.79) -- cycle ;
\draw    (330.14,240) -- (330.14,260) ;
\draw  [fill={rgb, 255:red, 28; green, 27; blue, 27 }  ,fill opacity=1 ] (332.98,260.5) .. controls (332.98,259.12) and (331.74,258) .. (330.23,258) .. controls (328.71,258) and (327.48,259.12) .. (327.48,260.5) .. controls (327.48,261.88) and (328.71,263) .. (330.23,263) .. controls (331.74,263) and (332.98,261.88) .. (332.98,260.5) -- cycle ;
\draw    (90,120) -- (90,100) ;
\draw  [fill={rgb, 255:red, 28; green, 27; blue, 27 }  ,fill opacity=1 ] (93,100.5) .. controls (93,99.12) and (91.77,98) .. (90.25,98) .. controls (88.73,98) and (87.5,99.12) .. (87.5,100.5) .. controls (87.5,101.88) and (88.73,103) .. (90.25,103) .. controls (91.77,103) and (93,101.88) .. (93,100.5) -- cycle ;

\draw    (170.14,120) -- (170.14,100) ;
\draw  [fill={rgb, 255:red, 28; green, 27; blue, 27 }  ,fill opacity=1 ] (173.14,100.5) .. controls (173.14,99.12) and (171.91,98) .. (170.39,98) .. controls (168.87,98) and (167.64,99.12) .. (167.64,100.5) .. controls (167.64,101.88) and (168.87,103) .. (170.39,103) .. controls (171.91,103) and (173.14,101.88) .. (173.14,100.5) -- cycle ;

\draw    (330.14,120) -- (330.14,100) ;
\draw  [fill={rgb, 255:red, 28; green, 27; blue, 27 }  ,fill opacity=1 ] (333.14,100.5) .. controls (333.14,99.12) and (331.91,98) .. (330.39,98) .. controls (328.87,98) and (327.64,99.12) .. (327.64,100.5) .. controls (327.64,101.88) and (328.87,103) .. (330.39,103) .. controls (331.91,103) and (333.14,101.88) .. (333.14,100.5) -- cycle ;

\draw  [dash pattern={on 4.5pt off 4.5pt}] (60,70) -- (366,70) -- (366,107.25) -- (60,107.25) -- cycle ;

\draw (91,140) node    {$\mathbf{x}_{1}$};
\draw (171,140) node    {$\mathbf{x}_{2}$};
\draw (331,140) node    {$\mathbf{x}_{K}$};
\draw (91,220) node    {$\mathbf{\Upsilon }_{1}$};
\draw (171,220) node    {$\mathbf{\Upsilon }_{2}$};
\draw (331,220) node    {$\mathbf{\Upsilon }_{K}$};
\draw (128.83,123) node    {$\phi ^{p}_{\mathbf{x}_{1,2} |\mathbf{Q}_{c}}$};
\draw (66.1,176.5) node    {$\phi ^{m}_{\mathbf{x}_{1} |\mathbf{\Upsilon }_{1}}$};
\draw (145.4,176.5) node    {$\phi ^{m}_{\mathbf{x}_{2} |\mathbf{\Upsilon }_{2}}$};
\draw (304.04,176.5) node    {$\phi ^{m}_{\mathbf{x}_{K} |\mathbf{\Upsilon }_{K}}$};
\draw (66.1,258.47) node    {$\phi ^{w}_{\mathbf{\Upsilon }_{1} |\boldsymbol{\Psi }}$};
\draw (145.4,258.47) node    {$\phi ^{w}_{\mathbf{\Upsilon }_{2} |\boldsymbol{\Psi }}$};
\draw (304.04,258.47) node    {$\phi ^{w}_{\mathbf{\Upsilon }_{K} |\boldsymbol{\Psi }}$};
\draw (94.36,84.36) node    {$\phi ^{m}_{\mathbf{x}_{1} |\mathbf{W}_{gt}}$};
\draw (174.5,84.36) node    {$\phi ^{m}_{\mathbf{x}_{2} |\mathbf{W}_{gt}}$};
\draw (334.5,84.36) node    {$\phi ^{m}_{\mathbf{x}_{K} |\mathbf{W}_{gt}}$};

\end{tikzpicture}
	\vspace{-2mm}
	\caption{\footnotesize Factor graph for our vehicle estimation problem in Experiment A (see Section \ref{sec:expA}). White circles represent random variables to be estimated (vehicle state $\mbf{x}$ and measurement covariances $\bs{\Upsilon}$). Small black dots represent factors in the joint likelihood of the data and the state. Binary motion prior factors, $\phi^p_{\mbf{x}_{k-1,k}|\mbf{Q}_c}$, depend on parameter $\mbf{Q}_c$. Unary groundtruth pose factors (if available), $\phi^m_{\mbf{x}_k | \mbf{W}_{gt}}$, depend on parameter $\mbf{W}_{gt}$.  Factors $\phi^m_{\mbf{x}_k|\mbf{\Upsilon}_k}$ and $\phi^w_{\mbf{\Upsilon}_k| \mbs{\Psi}}$ are for applying an Inverse-Wishart prior over our measurement pose covariances,  $\bs{\Upsilon}$, and depend on parameter $\bs{\Psi}$. We are able to learn parameters $\mbf{Q}_c$ and $\bs{\Psi}$, even without groundtruth factors (factors inside dashed box).}
	\label{fig:factor_graph}
	\vspace{-6mm}
\end{figure}
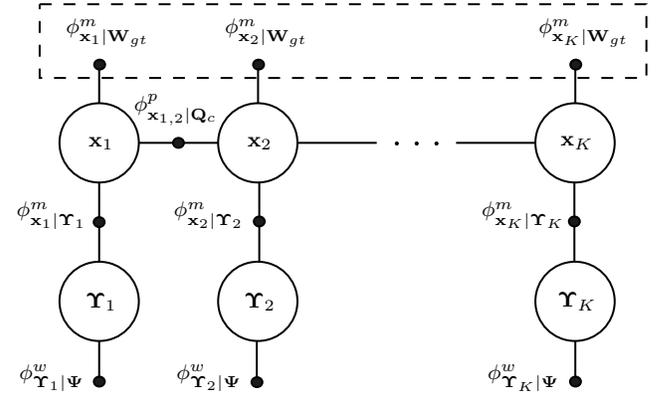

In this paper, we show that we can learn the parameters of a nonlinear system in concert with a nonlinear batch state estimation framework, namely \ac{ESGVI} \cite{barfoot_19}.   \ac{ESGVI} exploits the fact that the joint likelihood between the observed measurements (data) and the latent state can be factored, which provides a family of scalable state estimation tools starting from a variational inference objective.  To extend this to parameter learning, we use \ac{EM}. 
In the E-step, we fix all model parameters and optimize a bound on the data log-likelihood, the so-called \ac{ELBO}; this is equivalent to \ac{ESGVI} latent state inference. In the M-step, we hold the latent state estimate fixed and optimize the \ac{ELBO} for the parameters. Our method is general and applicable to any nonlinear system identification problem, even when the factorization of the joint likelihood has cycles (e.g., \ac{SLAM}). Barfoot et al. \cite{barfoot_19} hint at the \ac{ESGVI} extension to parameter learning, but do not demonstrate it in practice.

Our demonstration of parameter learning focuses on robot noise models. The noise models of the motion prior and observed measurements are often assumed to be known or tuned by trial and error. Our previous work demonstrated parameter learning for vehicle motion priors, but required accurate and complete (i.e., observation of the complete latent state) groundtruth \cite{wong2020data}. However, often times, collecting such groundtruth is not possible or extremely expensive. We demonstrate the ability to learn these noise models from only noisy measurements. If groundtruth is available, we treat it simply as another (noisy) measurement that can be included in the framework. We also demonstrate that an \ac{IW} prior over the time-varying measurement covariances, using a \ac{MAP} treatment in the variational setting, achieves outlier rejection in both parameter learning and latent state inference. 
We then demonstrate our parameter learning method on a real-world lidar dataset and a pose graph optimization problem created from a front-end pose graph SLAM algorithm. We show that our parameter learning method is able to handle both noisy measurements and outliers during training and testing.

In summary, the main contribution of this paper is a detailed investigation and experimental demonstration of parameter learning as part of ESGVI. Our application focuses on trajectory estimation, where we show nonlinear system identification using noisy measurements, without groundtruth. We also include outlier rejection in the variational setting by placing an \ac{IW} prior over covariances.

In Section \ref{sec:related_work} we review previous work. An overview of the ESGVI framework with parameter learning is provided in \ref{sec:background}. Section \ref{sec:methodology} presents the noise models we use and how we learn their parameters. An experimental evaluation of our parameter learning method is presented in \ref{sec:experiments}. In Section \ref{sec:conclusion}, we provide concluding remarks and discuss future work.

\section{Related Work} \label{sec:related_work}
System identification has been an active research area for decades \cite{aastrom1971system,gevers2006personal,kerschen2006past,fu2013research}.  In the interest of space, we restrict our review of related work to the techniques that are most similar to our proposed approach. In the domain of parameter learning, the most common approach is to find parameters that maximize the likelihood of the data. One way to do this is to directly maximize the likelihood function with respect to the parameters \cite{wong2020data,kokkala2014expectation,kokkala16}. This can be a difficult problem to solve, particularly when the model depends on missing or unobserved variables. In this case, an indirect approach can be taken by introducing a latent state to the problem, which can be estimated alongside of the parameters. This is known as Expectation Maximization (EM), an iterative algorithm that alternates between optimizing for a distribution over the latent state and the parameters.

Past work has shown how to estimate all the parameters of a linear dynamical system using EM, with Kalman smoothing in the E-step to update states and calculating analytic equations for parameter updates in the M-step \cite{shumway82}. There have also been methods that attempt parameter learning for nonlinear systems with EM. Ghahramani and Roweis \cite{ghahramani99} learn a full nonlinear model using Gaussian Radial Basis Functions (RBFs) to approximate the nonlinear expectations that would otherwise be intractable to compute. This method was applied to learn a simple sigmoid nonlinearity. Other methods approximate the required expectation using particle smoothing \cite{schon2011system} or sigmapoint smoothing \cite{kokkala2014expectation,kokkala16,gavsperin2011application}. These methods, however, did not learn a full nonlinear model, but only learned parameters of a mostly predefined model (e.g., calibration parameters), and were tested only in simulation.

Unlike all these other methods, we use ESGVI within the EM parameter learning framework, which is a more general method not limited to problems with a specific factorization of the joint likelihood between the data and the latent state (e.g., smoothing problems with a block-tridiagonal inverse covariance). We also demonstrate a practical application of parameter learning by estimating the parameters of our motion prior and measurement noise models in a batch estimation framework.

While we are interested in batch estimation, previous work has investigated learning the noise model parameters of filters. Abbeel et al. \cite{abbeel2005discriminative} learn the noise model parameters of the Kalman Filter offline. However, these parameters are assumed to be static and do not vary with time. One popular area of study that handles changing covariances is Adaptive Kalman Filtering, where the measurement covariance is updated in an online fashion based on the statistics of the measurement innovation \cite{mohamed1999adaptive,hu2003adaptive,yang2006optimal}. The measurement covariance in these cases is updated based solely on the data seen during inference, whereas we incorporate a prior.

Ko and Fox \cite{ko11} apply Gaussian process regression to learn robot measurement and motion models, but because they do not exploit sparsity, need to resort to using their learned models in a filter estimation framework. We exploit sparsity for batch estimation.
Recent methods take advantage of deep neural networks (DNNs) to learn the robot noise models \cite{brossard2019ai,liu2018deep,russell2019multivariate} but in many cases require groundtruth to train the DNN. We bypass this requirement by simultaneously estimating a distribution over the latent state.

Barfoot et al. \cite{barfoot_19} show how to learn a constant covariance using ESGVI through EM but do not demonstrate it in practice.
Our main contributions compared to \cite{barfoot_19} is demonstrating parameter learning for a specific application and learning time-varying covariances by introducing an IW prior over our covariances, which enables outlier rejection.
As an alternate method for outlier rejection, Chebrolu et al. \cite{chebrolu2020adaptive} use EM to learn a tuning parameter for M-estimation but treat their latent variables as point estimates.
The IW distribution has been used as a prior over covariances before, but the parameters were assumed to be known \cite{wilson2011generalised}. We seek to learn at least some of the parameters of the prior.

To the best of our knowledge, the work we present in this paper is the first attempt at wrapping parameter learning into the ESGVI framework to solve a practical problem, and shows that we can achieve a robust extension of ESGVI (with an outlier rejection scheme) by placing an IW prior on our measurement covariances. We also show comparable trajectory estimation performance between learning parameters with and without groundtruth.
\section{ESGVI with Parameter Learning} \label{sec:background}
\subsection{Variational Setup}
We begin with the maximum-likelihood problem for the given data, $\mbf{z}$, which is expressed as
\begin{equation} 
  \mbs{\theta}^\star = \arg \max_{\mbs{\theta}} p(\mbf{z} | \mbs{\theta}),  
\end{equation}
where $\mbs{\theta}$ represents the parameters of our system that we wish to learn.

We define the loss that we wish to minimize as the negative log-likelihood of the data and introduce the latent state, $\mbf{x}$. Applying the usual EM decomposition results in
\begin{align} \nonumber \label{eq:em_decomp}
  \mathscr{L} &= -\ln{p(\mbf{z} | \mbs{\theta})} = \underbrace{\int q(\mbf{x}) \ln \left( \frac{p(\mbf{x} | \mbf{z}, \mbs{\theta})}{q(\mbf{x})} \right) \, d\mbf{x}}_{\mbox{$\leq$ 0}} \\  &\qquad \qquad  \underbrace{-\int q(\mbf{x}) \ln \left( \frac{p(\mbf{x}, \mbf{z}  | \mbs{\theta})}{q(\mbf{x})} \right) \, d\mbf{x}}_{\mbox{upper bound}},
\end{align}
where we define our approximate posterior as a multivariate Gaussian distribution, $q(\mbf{x}) = \mathcal{N}(\mbs{\mu}, \mbs{\Sigma})$. We now proceed iteratively in two steps, the expectation step (E-step) and the maximization step (M-step)\footnote{We are working with the negative log-likelihood so we are technically applying Expectation Minimization, but the acronym stays the same.}.

As commonly done in the EM framework, in both the E-step and the M-step, we optimize the upper bound term in (\ref{eq:em_decomp}), which is also known as the (negative) Evidence Lower Bound (ELBO). Using the expression for the entropy, $-\int q(\mbf{x}) \ln q(\mbf{x}) d\mbf{x}$, for a Gaussian and dropping constants, the upper bound term is written as the loss functional of \ac{ESGVI},
\begin{equation}
\label{eq:functional}
V(q|\mbs{\theta}) = \mathbb{E}_q[ \phi(\mbf{x}|\mbs{\theta})] + \frac{1}{2} \ln \left( |\mbs{\Sigma}^{-1}| \right),
\end{equation}
where we define $\phi(\mbf{x}|\mbs{\theta}) = - \ln p(\mbf{x},\mbf{z}|\mbs{\theta})$, $\mathbb{E}_q[\cdot]$ is the expectation conditioned on the distribution $q(\mbf{x})$, and $|\cdot|$ is the matrix determinant. We drop $\mbf{z}$ in the notation for convenience as our expectation is over $\mbf{x}$.

Taking the derivatives of the loss functional with respect to $\mbs{\mu}$ and $\mbf{\Sigma}^{-1}$, Barfoot et al. \cite{barfoot_19} developed a Newton-style iterative optimizer to update our estimate of $q(\mbf{x})$. We summarize the optimization scheme here as
\begin{subequations}\label{eq:iterstein}
\begin{align}
\left(\mbs{\Sigma}^{-1}\right)^{(i+1)} & =   \sum_{k=1}^K \mbf{P}_k^T\mathbb{E}_{q^{(i)}_k} \left[ \frac{\partial^2  \phi_k( \mbf{x}_k |\mbs{\theta})}{\partial \mbf{x}_k^T \partial \mbf{x}_k}\right] \mbf{P}_k, \label{eq:itersteina}\\ 
\left(\mbs{\Sigma}^{-1}\right)^{(i+1)} \, \delta\mbs{\mu} & = - \sum_{k=1}^K \mbf{P}_k^T\mathbb{E}_{q^{(i)}_k} \left[ \frac{\partial \phi_k( \mbf{x}_k |\mbs{\theta})}{\partial \mbf{x}_k^T} \right], \label{eq:itersteinb}\\
\mbs{\mu}^{(i+1)} & = \mbs{\mu}^{(i)} + \delta\mbs{\mu}, \label{eq:itersteinc}
\end{align}
\end{subequations}
where superscript $i$ is used to denote variables at the $i^{\textrm{th}}$ iteration. We have exploited the factorization of the joint log-likelihood into $K$ factors as
\begin{equation}
  \phi(\mbf{x}|\mbs{\theta}) = \sum_{k=1}^K  \phi_k(\mbf{x}_k| \mbs{\theta}).
\end{equation}
For generality we have each factor, $\phi_k$, affected by the entire parameter set, $\mbs{\theta}$, but in practice it can be a subset. $\mbf{P}_k$ is a projection matrix that extracts $\mbf{x}_k$ from $\mbf{x}$
(i.e. $
  \mbf{x}_k = \mbf{P}_k\,\mbf{x}
$).
The marginal of $q$ associated with $\mbf{x}_k$ is
\begin{equation}
q_k(\mbf{x}_k) = \mathcal{N}(\mbf{P}_k \mbs{\mu}, \mbf{P}_k \mbs{\Sigma} \mbf{P}_k^T).
\end{equation}
Critical to the efficiency of the \ac{ESGVI} framework is the ability to compute the required marginals in (\ref{eq:itersteina}) and (\ref{eq:itersteinb}), without ever constructing the complete (dense) covariance matrix, $\mbs{\Sigma}$.  A sparse solver based on the method of Takahashi et al. \cite{takahashi73} is used to achieve this in \cite{barfoot_19}.

The expectations in (\ref{eq:itersteina}) and (\ref{eq:itersteinb}) can be approximated using Gaussian cubature samples (e.g., sigmapoints) of the marginal posterior. Importantly, approximating the expectations at only the mean of the posterior is equivalent to the \ac{MAP} batch optimization with Newton's method. Barfoot et al. \cite{barfoot_19} also provide a derivative-free optimization scheme with only Gaussian cubature, which we do not show here.

In the M-step, we hold $q(\mbf{x})$ fixed and optimize the upper bound for the parameters, $\mbs{\theta}$. We can optimize for $\mbs{\theta}$ by taking the derivative of the loss functional as follows:
\begin{align} \label{eq:deriv_loss} \nonumber
\frac{\partial V(q|\mbs{\theta})}{\partial \mbs{\theta}} &= \frac{\partial}{\partial \mbs{\theta}} \mathbb{E}_q[ \phi(\mbf{x}| \mbs{\theta})] = \frac{\partial}{\partial \mbs{\theta}} \mathbb{E}_q\left[ \sum_{k=1}^K  \phi_k(\mbf{x}_k| \mbs{\theta}) \right] \\ &= \sum_{k=1}^K \mathbb{E}_{q_k}\left[  \frac{\partial}{\partial \mbs{\theta}} \phi_k(\mbf{x}_k| \mbs{\theta}) \right].
\end{align}
In the last step, the expectation reduces from being over the full Gaussian, $q$, to the marginal associated with the variables in each factor, $q_k$.
We can then set the derivative to zero and isolate $\mbs{\theta}$ for a critical point, formulating an M-step. If isolation is not possible, we can use the gradient in (\ref{eq:deriv_loss}) for a partial M-step, which is known as \ac{GEM} \cite{bishop06}.

\subsection{Alternate Loss Functional}
In the E-step, we hold $\mbs{\theta}$ fixed and optimize $q(\mbf{x})$ for the best possible Gaussian fit. Barfoot et al. \cite{barfoot_19} present an alternate, Gauss-Newton-style loss functional for when the negative log-likelihood takes the form
\begin{equation} \label{eq:squared_mah_loss}
\phi(\mbf{x}|\mbf{W}) = \frac{1}{2} \left( \mbf{e}(\mbf{x})^T \mbf{W}^{-1}  \mbf{e}(\mbf{x}) - \ln(|\mbf{W}^{-1}|) \right),
\end{equation}
where $\mbs{\theta}$ is now a covariance matrix, $\mbf{W}$. With Jensen's inequality \cite{jensen06} and dropping the second term since $\mbf{W}$ is a constant in the E-step, we can write 
\begin{equation}
\mathbb{E}_q [ \mbf{e}(\mbf{x})]^T \mbf{W}^{-1} \mathbb{E}_q [ \mbf{e}(\mbf{x}) ] \leq \mathbb{E}_q \left[ \mbf{e}(\mbf{x})^T \mbf{W}^{-1}  \mbf{e}(\mbf{x}) \right].
\end{equation}
Motivated by this relationship, we can define a new loss functional for the E-step as
\begin{equation}
V^\prime(q) = \frac{1}{2} \mathbb{E}_q [ \mbf{e}(\mbf{x})]^T \mbf{W}^{-1} \mathbb{E}_q [ \mbf{e}(\mbf{x}) ] + \frac{1}{2} \ln( | \mbs{\Sigma}^{-1}| ),
\end{equation}
which is a conservative approximation of $V(q)$, appropriate for mild nonlinearities and/or concentrated posteriors. The alternate loss functional is simpler to implement in practice as it does not require the second derivative of the factors\footnote{Another alternative is the derivative-free optimization scheme with cubature sampling, at the cost of requiring more cubature samples
(i.e., more computation).
A derivative-free scheme for the alternate loss functional is also possible \cite{barfoot_19}.}. Also note how evaluating the expectation only at the mean of the posterior is equivalent to \ac{MAP} Gauss-Newton.
\section{Parameter Learning for Robot Noise Models} \label{sec:methodology}
\subsection{Constant Covariance} \label{sec:static_cov}
Barfoot et al. \cite{barfoot_19} outline parameter learning (M-step) for constant covariance noise models, which we summarize here. Our loss functional is
\begin{equation} \label{eq:loss_with_const_cov}
  V(q|\mbf{W}) = \mathbb{E}_q[ \phi^m(\mbf{x}|\mbf{W})] + \frac{1}{2} \ln \left( |\mbs{\Sigma}^{-1}| \right).
\end{equation}
This expression is similar to (\ref{eq:squared_mah_loss}), but we can exploit the factorization of $\phi^m(\mbf{x}|\mbf{W})$ to write:
\begin{align}
\label{eq:static_cov_factor}
 \phi^m(\mbf{x} | \mbf{W}) & =  \sum_{k=1}^K  \phi_k^{m}(\mbf{x}_k|\mbf{W})  \\
 & =  \sum_{k=1}^K \frac{1}{2}\left( \mbf{e}_k(\mbf{x}_k)^T \mbf{W}^{-1} \mbf{e}_k(\mbf{x}_k) - \ln(|\mbf{W}^{-1}|) \right), \nonumber
\end{align}
where the $K$ factors (in practice, it could be a subset) are affected by the unknown parameter $\mbf{W}$, a constant covariance matrix. Evaluating the derivative, as shown in (\ref{eq:deriv_loss}), with respect to $\mbf{W}^{-1}$ and setting to zero for computing a minimum results in the optimal $\mbf{W}$ to be
\begin{equation} \label{eq:static_cov}
\mbf{W} = \frac{1}{K} \sum_{k=1}^K \mathbb{E}_{q_k}\left[  \mbf{e}_k(\mbf{x}_k)  \mbf{e}_k(\mbf{x}_k)^T \right],
\end{equation}
which can be approximated with Gaussian cubature if the error functions, $\mbf{e}_k(\mbf{x}_k)$, are nonlinear. 

Alternatively, we can choose to linearize $\mbf{e}_k(\mbf{x}_k)$ at the posterior marginal, $q_k = \mathcal{N}(\mbs{\mu}_k, \mbs{\Sigma}_k)$, resulting in the following M-step:
\begin{align} \nonumber
  \mbf{W} &\approx \frac{1}{K} \sum_{k=1}^{K} \mathbb{E}_{q_k}\left[ \left( \mbf{e}_k(\mbs{\mu}_k) + \mbf{E}_k\, \delta\mbf{x}_k \right) \left( \mbf{e}_k(\mbs{\mu}_k) + \mbf{E}_k\, \delta\mbf{x}_k \right)^T \right] \\
  &= \frac{1}{K} \sum_{k=1}^{K} \left(\mbf{e}_k(\mbs{\mu}_k)\mbf{e}_k^T(\mbs{\mu}_k) + \mbf{E}_k \mbf{\Sigma}_k\mbf{E}_k^T\right), \label{eq:static_cov_lin}
\end{align}
where
$
  \mbf{E}_k = \left.\frac{\partial \mbf{e}_k(\mbf{x}_k)}{\partial \mbf{x}_k} \right|_{\mbf{x}_k = \mbs{\mu}_k}. 
$

\subsection{White-Noise-On-Acceleration Prior on Latent State}
We next demonstrate parameter learning for the situation where the covariance of a factor is indirectly estimated through another quantity. Consider the example of a \ac{WNOA} motion prior on the latent state, where the parameter we wish to estimate is the power-spectral density matrix, $\mbf{Q}_c$ \cite{barfoot2014batch}. More specifically, let us study the application of the prior in $SE(3)$ \cite{anderson2015full}, which is defined as follows:
\begin{equation} \label{eq:se3_prior}
\begin{split}
\dot{\mathbf{T}}(t)&=\bs{\varpi}(t)^\wedge{}\mathbf{T}(t), \\
\dot{\bs{\varpi}}&=\mathbf{w}(t),\quad \mathbf{w}(t) \sim \mathcal{GP}(\mathbf{0}, \mathbf{Q}_c\delta(t-t')),
\end{split}
\end{equation}
where \(\mathbf{T}(t) \in SE(3)\) is the pose expressed as a transformation matrix, \(\boldsymbol{\varpi}(t) \in \Real{^6}\) is the body-centric velocity, \(\mathbf{w}(t) \in \Real{^6}\) is a zero-mean, white-noise Gaussian process, and the operator, $\wedge$, transforms an element of $\Real{^6}$ into a member of Lie algebra, $\mathfrak{se}(3)$. The state at time $t_k$ is $\mathbf{x}_k = \{ \mathbf{T}_k, \varpii_k \}$, and similarly, $\mathbf{x}_{k-1,k}$ is the state at two consecutive times, $t_{k-1}$ and $t_k$.

We express the factors of our loss functional from (\ref{eq:functional}), but with only \ac{WNOA} prior factors for simplicity:
\begin{align}\label{eq:wnoa_factor}
\phi^p(\mbf{x}|\mbf{Q}_c) &= \sum_{k=2}^{K} \phi_k^p(\mbf{x}_{k-1,k}|\mbf{Q}_c)\\&= \sum_{k=2}^{K} \frac{1}{2}\left( \mbf{e}_{p,k}^T\mbf{Q}_k^{-1}\mbf{e}_{p,k} +\ln\lvert\mbf{Q}_k\rvert \right), \nonumber
\end{align}
where
\begin{equation}\label{eq:wnoa_error}
\mbf{e}_{p,k} = \begin{bmatrix}
\ln(\Tsmall_k\Tsmall_{k-1}^{-1})^\vee-(t_k-t_{k-1})\bs{\varpi}_{k-1} \\
\jac^{-1}(\ln(\Tsmall_k\Tsmall_{k-1}^{-1})^\vee)\bs{\varpi}_k-\bs{\varpi}_{k-1}
\end{bmatrix},
\end{equation}
and the covariance of the prior, $\mbf{Q}_k$, is defined as \cite{barfoot2014batch}
\begin{equation*}
  \mbf{Q}_k=\mbf{Q}_{\Delta t}\otimes\mbf{Q}_c, \quad
  \mbf{Q}_k^{-1}=\mbf{Q}_{\Delta t}^{-1}\otimes\mbf{Q}_c^{-1},
\end{equation*}
\vspace{0.001mm}
\begin{equation*}
\mbf{Q}_{\Delta t}=\begin{bmatrix}
  \frac{1}{3}\Delta t^3 & \frac{1}{2}\Delta t^2 \\
  \frac{1}{2}\Delta t^2 & \Delta t
\end{bmatrix}, \quad
\mbf{Q}_{\Delta t}^{-1}=\begin{bmatrix}
  12\Delta t^{-3} & -6\Delta t^{-2} \\
  -6\Delta t^{-2} & 4\Delta t^{-1}
\end{bmatrix},
\end{equation*}
where $\otimes$ is the Kronecker product.
Solving for the derivative with respect to $Q_{c_{ij}}$, the $(i,j)$ matrix element of $\mbf{Q}_c$, we have
\begin{multline}
\frac{\partial V(q|\mbs{\theta})}{\partial Q_{c_{ij}}}=\frac{1}{2}\mbox{tr}\left(\sum_{k=2}^{K}{\mathbb{E}_{q_{k-1,k}}[\mbf{e}_{p,k}\mbf{e}_{p,k}^T](\mbf{Q}_{\Delta t}^{-1}\otimes\mbf{1}_{ij})}\right) \\
-\frac{1}{2}(K-1)\textrm{dim}(\mbf{Q}_{\Delta t})Q_{c_{ij}},
\end{multline}
where $q_{k-1,k}$ is the marginal posterior at two consecutive times, $t_{k-1}$ and $t_k$. Setting the derivative to zero, the optimal estimate of our parameter is then
\beq \label{eq:wnoa}
Q_{c_{ij}}=\frac{\mbox{tr}\left(\sum_{k=2}^{K}{\mathbb{E}_{q_{k-1,k}}[\mbf{e}_{p,k}\mbf{e}_{p,k}^T](\mbf{Q}_{\Delta t}^{-1}\otimes\mbf{1}_{ij})}\right)}{\textrm{dim}(\mbf{Q}_{\Delta t})(K-1)}.
\eeq
As explained for (\ref{eq:static_cov}), the expectation in (\ref{eq:wnoa}) can be approximated with Gaussian cubature or linearization.

\subsection{Inverse-Wishart Prior on Covariance} \label{sec:iw_prior}
We further extend covariance estimation by incorporating a prior. Instead of treating the covariance as a static parameter, we treat it as a random variable and place an IW prior on it. We then learn some of the parameters of the prior. In order to do so, we redefine our joint likelihood as
\begin{equation}
  p(\mbf{x},\mbf{z},\mbf{R}) = p(\mbf{x},\mbf{z}|\mbf{R})p({\mbf{R}}),
\end{equation}
where now we also include the covariances, $\mbf{R} =\{ {\mbf{R}_1, \mbf{R}_2, \dots \mbf{R}_K}\}$, as random variables. We also redefine our posterior estimate to be
\begin{equation}
  q'(\mbf{x}) = q(\mbf{x})s(\mbf{R}),
\end{equation}
a product between a Gaussian $q(\mbf{x})$ and a posterior distribution for the covariances, $s(\mbf{R})$.

The upper bound term in the \ac{EM} decomposition of (\ref{eq:em_decomp}) can now be written as
\begin{equation}
  -\int\int q(\mbf{x})s(\mbf{R}) \ln{\left(\frac{p(\mbf{x},\mbf{z}|\mbf{R})p({\mbf{R}})}{q(\mbf{x})s(\mbf{R})}\right)} \,d\mbf{x}\,d\mbf{R}.
\end{equation}
We define the posterior over the covariances as
\begin{equation}
  s(\mbf{R}) = \delta(\mbf{R} - \mbf{\Upsilon}),
\end{equation}
where $\delta(\cdot)$ is the Dirac delta function (interpreted as a probability density function) and $\mbf{\Upsilon} = \{\mbf{\Upsilon}_1,\mbf{\Upsilon}_2 \dots \mbf{\Upsilon}_K \}$ is the set of optimal covariances. The upper bound now simplifies to
\begin{align} 
  -\int q(\mbf{x}) &\ln{\left(p(\mbf{x},\mbf{z}| \mbf{\Upsilon})p( \mbf{\Upsilon})\right)} \, d\mbf{x} \\ &+ \int q(\mbf{x}) \ln{q(\mbf{x})}\,d\mbf{x} + \underbrace{\int s(\mbf{R})\ln{s({\mbf{R}})}d\mbf{R}}_{\mbox{indep. of $\mbf{\Upsilon}$}}, \nonumber \end{align}
where we have abused notation and written $p(\mbf{R=\Upsilon})$ as $p(\mbf{\Upsilon})$, and similarly will later write $p(\mbf{R}_k=\mbf{\Upsilon}_k)$ as $p(\mbf{\Upsilon}_k)$. 
We view our selection of the delta function as a convenient way of showing how we can approximate a Gaussian distribution for the trajectory and a \ac{MAP} approximation of the covariances in a single variational framework. The last term is the differential entropy of a Dirac delta function, and because it is independent of our variational parameter, $\mbf{\Upsilon}$, we choose to drop it from our loss functional.

We assume $p(\mbf{\Upsilon})$ factors as
$
  p( \mbf{\Upsilon}) = \prod_{k=1}^K p(\mbf{\Upsilon}_k).
$
We apply an \ac{IW} prior over our covariances by defining
\begin{equation}
p(\mbf{\Upsilon}_k) =\frac{\vert \mbs{\Psi}\vert ^{\nu/2}}{2^{\frac{\nu d}{2}\Gamma_{d}(\frac{\nu}{2})}}\vert \mbf{\Upsilon}_k\vert ^{-\frac{\nu+d+1}{2}}\exp \left(- \frac{1}{2}\mbox{tr}(\mbs{\Psi} \mbf{\Upsilon}_k^{-1})\right),
\end{equation}
where $d$ is the dimension of $\mbf{\Upsilon}_k$, $\mbs{\Psi}\in\mathbb{R}^{d\times d} > 0$ is the scale matrix, $\nu>d-1$ is the degrees-of-freedom (DOF), and $\Gamma_{d}(\cdot)$ is the multivariate Gamma function. The \ac{IW} distribution has been used as a prior over covariance matrices before, which led to outlier rejection at inference \cite{barfoot2017state, peretroukhin2016probe}, but the parameters of the prior were assumed to be known. We choose to estimate the scale matrix parameter $\mbs{\Psi}$ and leave the degrees-of-freedom $\nu$ as a metaparameter.

Now we define our factors as
\begin{align} \nonumber\label{eq:iw}
  -\ln{(\mbf{\Upsilon})} &= \sum_{k=1}^K -\ln{p(\mbf{\Upsilon}_k)} \\ &= \sum_{k=1}^K \phi_k^w(\mbf{\Upsilon}_k|\mbf{\Psi}) = \phi^w(\mbf{\Upsilon}|\mbf{\Psi}).
\end{align}

Dropping constant terms, the loss functional can finally be written as
\begin{multline}
  V(q'|\mbf{\Upsilon},\bs{\Psi}) = \sum_{k=1}^K \mathbb{E}_{q_k}\left[ \phi_k^m(\mbf{x}_k|\mbf{\Upsilon}_k) + \phi_k^w(\mbf{\Upsilon}_k|\mbf{\Psi})\right] \\+ \frac{1}{2} \ln \left( |\mbs{\Sigma}^{-1}| \right),
\end{multline}
where
\begin{align}
  &\phi_k^m(\mbf{x}_k|\mbf{\Upsilon}_k) = \frac{1}{2} \left( \mbf{e}_k(\mbf{x}_k)^T \mbf{\Upsilon}_k^{-1} \mbf{e}_k(\mbf{x}_k) - \ln(|\mbf{\Upsilon}_k^{-1}| \right), \\
  &\phi_k^w(\mbf{\Upsilon}_k|\bs{\Psi}) = -\frac{\alpha-1}{2}\ln{|\mbf{\Upsilon}_k^{-1}|} - \frac{\nu}{2}\ln{|\mbs{\Psi}|} + \frac{1}{2} \mbox{tr}({\mbs{\Psi}} \mbf{\Upsilon}_k^{-1}),\label{eq:iw2}
\end{align}
with $\alpha=\nu+d+2$.

In the E-step, we hold $\mbs{\Psi}$ fixed and optimize for $\mbf{\Upsilon}_k$, which we accomplish by taking the derivative of the loss functional as follows:
\begin{equation}
  \frac{\partial V}{\partial \mbf{\Upsilon}_k^{-1}} = \frac{1}{2}\mathbb{E}_{q_k}\left[  \mbf{e}_k(\mbf{x}_k)  \mbf{e}_k(\mbf{x}_k)^T \right]-\frac{1}{2}\alpha \mbf{\Upsilon}_k+\frac{1}{2}\bs{\Psi}.
\end{equation}
Setting the derivative to zero,
\begin{align} \label{eq:optimal_rk}
  \mbf{\Upsilon}_k &= \frac{1}{\alpha}\mbs{\Psi}+\frac{1}{\alpha}\mathbb{E}_{q_k}\left[  \mbf{e}_k(\mbf{x}_k)  \mbf{e}_k(\mbf{x}_k)^T \right] \\
  & = \frac{\alpha-1}{\alpha} \underbrace{\left(\frac{\mbs{\Psi}}{\alpha-1}\right)}_{\mbox{\ac{IW} mode}} + \frac{1}{\alpha}\mathbb{E}_{q_k}\left[  \mbf{e}_k(\mbf{x}_k)  \mbf{e}_k(\mbf{x}_k)^T \right], \nonumber
\end{align}
where we see the optimal $\mbf{\Upsilon}_k$ is a weighted average between the mode of the \ac{IW} distribution and the optimal static covariance estimate from (\ref{eq:static_cov}) at a single marginal factor. Since our E-step in \ac{ESGVI} is already iterative, we can seamlessly extend it by applying (\ref{eq:optimal_rk}) as \ac{IRLS}.

In the M-step, we hold $\mbf{\Upsilon}$ fixed and optimize for $\mbs{\Psi}$, which we accomplish by taking the derivative of the loss functional as follows:
\begin{equation}
  \frac{\partial V}{\partial \mbs{\Psi}} = \sum_{k=1}^K \left(-\frac{\nu}{2}\bs{\Psi}^{-1}+\frac{1}{2}\mbf{R}_k^{-1} \right).
\end{equation}
Setting the derivative to zero,
\begin{equation} \label{eq:optimal_psi}
  \mbs{\Psi}^{-1} = \frac{1}{K\nu} \sum_{k=1}^K \mbf{\Upsilon}^{-1}_k.
\end{equation}
Applying (\ref{eq:optimal_rk}) in the E-step and (\ref{eq:optimal_psi}) in the M-step, we found that our optimzation scheme was still ill-posed, and our covariance estimates tended toward the positive-definite boundary (i.e., the zero matrix). We propose constraining the determinant of $\mbf{\Psi}$ to be a constant $\beta$, which can be thought of as constraining the volume of the uncertainty ellipsoid of the corresponding measurements to be fixed. We accomplish this by scaling the latest $\mbf{\Psi}$ update as follows:
\begin{equation}
  \mbf{\Psi}_{\text{constrained}} \leftarrow \left( \beta \, |\mbf{\Psi}|^{-1}\right)^{\frac{1}{d}} \, \mbf{\Psi}.
\end{equation}
We then rely on the noise models of other factors (e.g., the motion prior) to adapt to our selection of $\beta$ during training.

\section{Experimental Validation}\label{sec:experiments}
To evaluate our parameter learning method, we will be working with the vehicle dataset collected and used in our previous work \cite{wong2020data}. The dataset consists of 36 km of driving, with Velodyne VLS-128 lidar data and an Applanix POS-LV positioning system.
There are two sources of 6-DOF vehicle pose measurements. The first is from the POS-LV system, which we treat as groundtruth. The second is from a lidar localization system from Applanix, which localizes the lidar data to a prebuilt high-definition map.

We use Route A\footnote{\enspace Map available at: \url{https://tinyurl.com/rrjgxaj}}, our 16 km long training set, to learn the parameters of our noise models. For inference, we perform a batch trajectory optimization on Route B\footnote{\enspace Map available at: \url{https://tinyurl.com/r5m78nq}}, our 20 km long test set, using the learned noise model parameters of our motion prior and measurements.

Finally, we evaluate our method on a pose graph optimization problem with false loop closures.

\subsection{Training With and Without Groundtruth} \label{sec:expA}
In Experiment A, our first experiment, we only use the lidar localization measurements to learn our model parameters (training without groundtruth). As a benchmark, we also learn another set of model parameters where we additionally include groundtruth poses in our training (training with incomplete groundtruth). This is different from our previous work \cite{wong2020data} where the training method required groundtruth of the entire state (training with complete groundtruth), which for our problem setup is pose and body-centric velocity. Additionally, in that paper, the measurement covariances were assumed to be known and not learned.

The loss functional corresponding to this experiment is
\begin{multline}
V(q'|\mbf{\Upsilon},\mbs{\Psi},\mbf{W}_{gt}, \mbf{Q}_c)=  \,\mathbb{E}_{q'}[\phi^p(\mbf{x}|\mbf{Q}_c) + \phi^m(\mbf{x} | \mbf{W}_{gt})\\
+\phi^m(\mbf{x}|\mbf{\Upsilon})+\phi^w(\mbf{\Upsilon}|\bs{\Psi})]+
\frac{1}{2} \ln \left( |\mbs{\Sigma}^{-1}|\right),
\end{multline}
where $\phi^p(\mbf{x}|\mbf{Q}_c)$ are the WNOA prior factors, $\phi^m(\mbf{x} | \mbf{W}_{gt})$ are the groundtruth factors (when available), and $\phi^m(\mbf{x}|\mbf{\Upsilon})$ and $\phi^w(\mbf{\Upsilon}| \mbs{\Psi})$ are the lidar measurement factors with an IW prior over the covariances. See (\ref{eq:wnoa_factor}) for the definition of $\phi^p(\mbf{x}|\mbf{Q}_c)$ and (\ref{eq:static_cov_factor}) for the definition of $\phi^m(\mbf{x} | \mbf{W}_{gt})$ and $\phi^m(\mbf{x}|\mbf{\Upsilon})$. For the definition of $\phi^w(\mbf{\Upsilon}|\bs{\Psi})$, see (\ref{eq:iw}) and (\ref{eq:iw2}).

The WNOA error function (required for $\phi^p$) is shown in (\ref{eq:wnoa_error}), and the error function for pose measurements (required for $\phi^m$) is defined as
\begin{equation}
\mbf{e}_{m,k}=\ln(\Tsmall_k\Tsmall_{\textrm{meas},k}^{-1}).
\end{equation}

The estimation problem in this experiment can be represented by the factor graph in Figure \ref{fig:factor_graph}, where we can train with or without the groundtruth factors, which are shown inside the dashed box. For the sake of conciseness in our notation, we denote $\phi^p(\mbf{x}_{k-1,k}|\mbf{Q}_c)$ as $\phi^p_{\mbf{x}_{k-1,k}|\mbf{Q}_c}$, $\phi^m(\mbf{x}_k | \mbf{W}_{gt})$ as $\phi^m_{\mbf{x}_k | \mbf{W}_{gt}}$, $\phi^m(\mbf{x}_k|\mbf{\Upsilon}_k)$ as $\phi^m_{\mbf{x}_k|\mbf{\Upsilon}_k}$, and $\phi^w(\mbf{\Upsilon}_k| \mbs{\Psi})$ as $\phi^w_{\mbf{\Upsilon}_k| \mbs{\Psi}}$.

We choose to fix the parameters to $\nu=6$ and $\beta=1$ and learn the parameters $\bs{\Psi}$, $\mbf{W}_{gt}$ (when groundtruth is available), and $\mbf{Q}_c$. For both sets of learned parameters, we then perform trajectory estimation on our test set, where we only use the lidar localization measurements with our learned covariance model and our learned motion prior.

\begin{figure}[b]
	\centering
	\vspace{-7mm}
	\includegraphics[trim={0 0 0 3mm},clip]{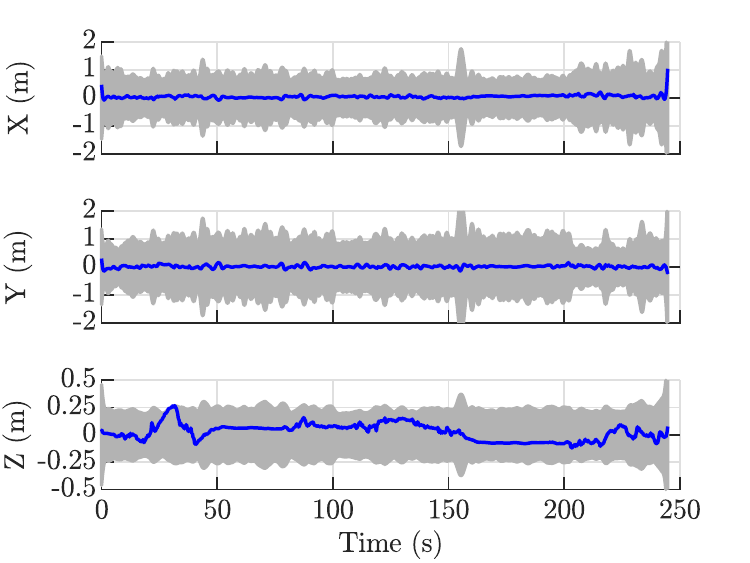}
	\vspace{-2mm}
	\caption{\footnotesize Experiment A - Error plots (blue lines) along with the $3\sigma$ covariance envelopes (gray background) when parameters are trained without groundtruth.
		}
	\label{fig:error_plot}
\end{figure}

Figure \ref{fig:error_plot} shows the error plots where we have trained without groundtruth for our estimated $x$, $y$, and $z$ positions, along with their $3\sigma$ covariance envelopes. As can be seen, the errors consistently remain within the covariance envelopes. We do however note that our estimator appears to be underconfident. We believe that this is a result of our decision to constrain $\lvert\bs{\Psi}\rvert=\beta=1$ in order for our training method to work in practice. This decision is analogous to fixing the volume of the covariance ellipsoid to be constant. In doing so, we relied on the learned covariance of the motion prior to adjust relative to the measurement covariances. The posterior mean is unaffected by this choice but not the posterior covariance.
\begin{table}[!t]
	\vspace{1mm}
	\centering
	\caption{\footnotesize \label{tab:compare_to_gt} Experiment A - Comparison of translational errors on test set between training with complete groundtruth, with incomplete groundtruth, and without groundtruth (GT). We note that the first column, our previous work, did not learn the measurement covariances.}
	\begin{tabular}{cccc}
		\begin{tabular}[c]{@{}c@{}}Seq \\ no.\end{tabular} &
		\begin{tabular}[c]{@{}c@{}}Trained with \\
			complete GT \textsuperscript{\cite{wong2020data}}\\ (m)\end{tabular} &
		\begin{tabular}[c]{@{}c@{}}Trained with \\ incomplete GT \\ (m)\end{tabular} &
		\begin{tabular}[c]{@{}c@{}}Trained \\ without GT \\ (m)\end{tabular} \\
		\hline
		0 & 0.0690 & 0.0720 &	0.0717 \\
		1 & 0.0888 & 0.1003 &	0.0925 \\
		2 & 0.4071 & 0.4148 &	0.4106 \\
		3 & 0.1947 & 0.1908 &	0.1847 \\
		4 & 0.2868 & 0.2866 &	0.2820 \\
		5 & 0.5703 & 0.5592 &	0.5549 \\
		6 & 0.3292 & 0.3014 &	0.2965 \\
		7 & 0.2207 & 0.2248 &	0.2230 \\
		8 & 0.1115 & 0.1151 &	0.1199 \\
		9 & 0.0979 & 0.1026 &	0.0997 \\
		\hline
		overall & 0.2376 & 0.2368 &	0.2335
	\end{tabular}
	\vspace{-4mm}
\end{table}

Table \ref{tab:compare_to_gt} shows the resulting mean translational errors from both training methods on all test sequences. We also include the results from our previous work where we trained using complete groundtruth for comparison.

While we achieve very similar errors across all training methods, the benefit is that we now do not require any groundtruth. Neither of the three training methods seem to outperform the others. We believe this is because our lidar localization measurements are quite accurate relative to groundtruth \cite{wong2020data}.

To further validate our method and show that we can indeed train with noisy measurements, we decided to artificially add additional noise to the measurements, where the noise statistics are unknown to the training process. We use the following $SE(3)$ perturbation scheme \cite{barfoot2017state,barfoot2014associating} to inject noise into the position portion of our pose measurements:
\begin{equation}
\Tsmall_{\textrm{noisy}}=\exp(\bs{\xi}^\wedge)\Tsmall_{\textrm{meas}},
\end{equation}
where
$
\bs{\xi}=\begin{bmatrix}
\bs{\xi}_{1:3} \\ \mbf{0}
\end{bmatrix},\quad
\bs{\xi}_{1:3}\sim\mathcal{N}\left(\mbf{0},
\sigma^2\mbf{I}
\right).
$

We vary $\sigma$ from 0.25 m to 1 m, injecting the same amount of noise into the test measurements and training measurements.
\begin{table}[!t]
	\vspace{1mm}
	\centering
	\caption{\footnotesize \label{tab:noise_exp} Experiment A - Analysis of how increasing noise on measurements affects the parameter learning method. Even with measurement errors of over 1.6 m, the errors on the estimated trajectory are under 0.5 m.}
	\vspace{-1mm}
	\begin{tabular}{c|c}
		\begin{tabular}[c]{@{}c@{}}Measurement errors (m)\end{tabular} &
		\begin{tabular}[c]{@{}c@{}}Estimated trajectory errors (m)\end{tabular} \\
		\hline
		0.2407 &	0.2335 \\
		0.5010 &	0.2909 \\
		0.8653 &	0.3289 \\
		1.2481 &	0.3936 \\
		1.6383 &	0.4566
	\end{tabular}
	\vspace{-4mm}
\end{table}

Table \ref{tab:noise_exp} shows how our test errors change with increasing noise on measurements in both our training and test set.
While measurement error increases significantly, up to over 1.6 m, we are still able to achieve translational errors of below 0.5 m on our estimated trajectory. This shows that we are still able to learn reasonable parameters of our system even without any groundtruth and quite noisy measurements.

\vspace{-5mm}
\subsection{Training and Testing With Measurement Outliers} \label{sec:expB}
In Experiment B, we show that estimating time-varying covariances for each of our measurements with an IW prior results in outlier rejection.
We artificially introduce outliers in our training and test set using the following method. With 5\% probability, we apply the following perturbation to the actual pose measurement:
\begin{equation}
\mbf{T}_{\textrm{outlier}}=\exp(\bs{\xi}^\wedge)\mbf{T}_{\textrm{meas}},
\end{equation}
with $\bs{\xi}\in\Real^{6}\sim \mathcal{U}(-200,200)$.

\begin{figure}[b] 
	\vspace{-7mm}
	\centering
	\subfigure[Measurement outliers]{%
		\includegraphics[height=4cm]{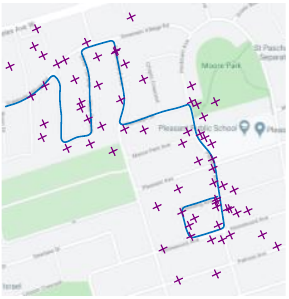} \label{fig:outliers}
	} 
	\quad 
	\subfigure[Concentration of errors]{%
		\includegraphics{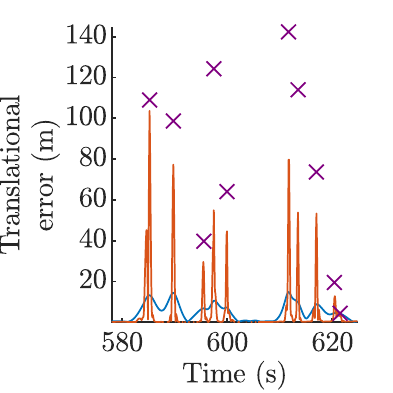} \label{fig:staticR_errors}
	}
	\vspace{-2mm}
	\caption{Experiments B \& C - \textbf{(a)} Measurement outliers (purple) overlaid with the groundtruth trajectory (blue). \textbf{(b)} Translational errors for the static covariance method on a portion of the test set containing measurement outliers (purple) when training with and without outliers (Experiments B in blue and C in orange, respectively).}
	\vspace{-6mm}
\end{figure}

Figure \ref{fig:outliers} shows an example of the measurement outliers on sequence 3 of our test set.

We now seek to compare the performance between the cases where we have treated the measurement covariance, $\mbf{W}$, as a static parameter to be learned, and where we have treated the measurement covariance at each time as a random variable and learn the parameter, $\bs{\Psi}$, of the IW prior.

The loss functional corresponding to the static measurement covariance is
\begin{multline}
V(q'|\mbf{W}, \mbf{Q}_c)=  \,\mathbb{E}_{q'}[\phi^p(\mbf{x}|\mbf{Q}_c) + \phi^m(\mbf{x} | \mbf{W})]\\
+\frac{1}{2} \ln \left( |\mbs{\Sigma}^{-1}|\right),
\end{multline}
whereas for the IW prior on the measurement covariances, the loss functional is
\begin{multline}
V(q'|\mbf{\Upsilon},\mbs{\Psi}, \mbf{Q}_c)=  \,\mathbb{E}_{q'}[\phi^p(\mbf{x}|\mbf{Q}_c) + \phi^m(\mbf{x}|\mbf{\Upsilon})\\+\phi^w(\mbf{\Upsilon}|\bs{\Psi})]+
\frac{1}{2} \ln \left( |\mbs{\Sigma}^{-1}|\right).
\end{multline}

Table \ref{tab:test_with_outliers} shows the resulting translational errors on our test trajectory. We can see that without the IW prior, the estimation framework fails to reject outliers, resulting in an overall translation error of above 5 m. However, using the IW prior, we see that the error is only 0.2365 m. When we did not have any outliers at all in our data, the error was 0.2335 m (Table \ref{tab:compare_to_gt}), meaning the average translational error on our test set only increased by 0.003 m.

From this experiment, we can see that using the IW prior allows for the handling of outliers in both training and testing due to our ability to estimate measurement covariances.
\begin{table}[!t]
	\centering
	\caption{\footnotesize \label{tab:test_with_outliers} Experiments B \& C - Translational errors using a static measurement covariance compared to using an IW prior when we have outliers in our test set. In Experiment B, we train with outliers and in Experiment C, we train without outliers.}
	\vspace{-2mm}
	\begin{tabular}{c|cc|cc}	
		& \multicolumn{2}{c}{Experiment B} & \multicolumn{2}{|c}{Experiment C} \\
		\hline
		\begin{tabular}[c]{@{}c@{}}Seq \\ no.\end{tabular} &
		\begin{tabular}[c]{@{}c@{}}Static $\mbf{W}$ \\ (m)\end{tabular} &
		\begin{tabular}[c]{@{}c@{}}IW prior \\ (m)\end{tabular} &
		\begin{tabular}[c]{@{}c@{}}Static $\mbf{W}$ \\ (m)\end{tabular} &
		\begin{tabular}[c]{@{}c@{}}IW prior \\ (m)\end{tabular} \\
		\hline
		0 & 6.1976	& 0.0773	& 7.3504	& 0.0731 \\
		1 & 5.8371	& 0.0979	& 6.0754	& 0.0948 \\
		2 & 5.3652	& 0.4125	& 5.5771	& 0.4096 \\
		3 & 5.1217	& 0.1860	& 5.8157	& 0.1873 \\
		4 & 5.5186	& 0.2807	& 5.5503	& 0.2826 \\
		5 & 5.4780	& 0.5563	& 6.3057	& 0.5554 \\
		6 & 6.3936	& 0.3004	& 7.1858	& 0.2995 \\
		7 & 5.6898	& 0.2274	& 6.0332	& 0.2256 \\
		8 & 6.3717	& 0.1233	& 9.3079	& 0.1224 \\
		9 & 6.8032	& 0.1036	& 8.1237	& 0.1046 \\
		\hline
		overall & 5.8776	& 0.2365	& 6.7325	& 0.2355
	\end{tabular}
	\vspace{-5mm}
\end{table}

\vspace{-2mm}
\subsection{Training Without and Testing With Measurement Outliers} \label{sec:expC}
In Experiment B,  we included outlier measurements in both the training and test set and saw that the IW prior allows us to achieve comparable errors to the case with no outliers. To see if this still holds even when we do not see any outliers in training, in Experiment C we train without any outliers but test with outliers. As the only difference between Experiment B and Experiment C is that we now train without any outliers, the loss functionals remain the same.

Table \ref{tab:test_with_outliers} shows that the resulting translational errors are again very high when we simply learn a static measurement covariance, but that we can still achieve reasonably low errors when learning the parameters of our IW prior. By incorporating the IW prior instead of learning a static measurement covariance, we decrease error from above 6~m to 0.2355 m. Compared to the error of 0.2335 m when there are no outliers in our test set (Table \ref{tab:compare_to_gt}), we see an increase in error of only 0.002 m with the IW prior.
This experiment shows that we can indeed still benefit from the outlier rejection scheme that comes with using an IW prior even when there are no outliers in our training set.

Comparing the results from Experiments B and C, we see that in both cases, incorporating the IW prior helps to reject outliers in the test set, regardless of whether there were any outliers in the training set. While errors for the static measurement covariance were similarly poor in both experiments, we note that the concentration of the errors are different as shown in Figure \ref{fig:staticR_errors}. What we see is that when we train without any outliers (Experiment C), the errors are concentrated where the outliers are in the test set. This result is unsurprising given that no outliers were seen in training, which in turn is reflected in our learned noise model parameters. When we train with outliers (Experiment B), the errors still peak around the outliers, but are more spread out over the entire trajectory. Regardless of this difference, we can see that using the IW prior is robust to both cases and still results in low translational errors.


\vspace{-1.4mm}
\subsection{Bicocca Dataset} \label{sec:bicocca}
\begin{figure}[t] 
	\centering
	\subfigure[Loop closures]{%
		\includegraphics[width=3.5cm, trim={1mm  0  5mm 0},clip]{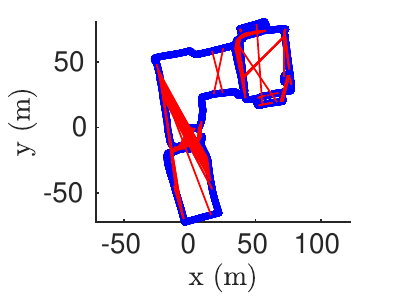} \label{fig:a}
	} 
	\quad 
	\subfigure[Optimzed pose graph]{%
		\includegraphics[width=3.5cm, trim={1mm  0  5mm 0},clip]{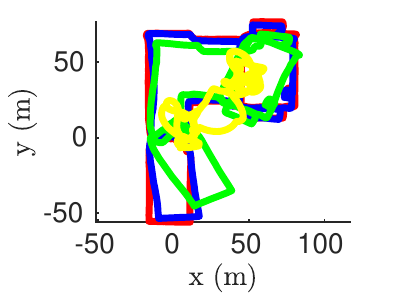} \label{fig:b}
	}
	\vspace{-2mm}
	\caption{\textbf{(a)} Odometric trajectory (blue) with loop closures (red). \textbf{(b)} Optimized trajectory using IW prior (blue), static covariance learning (green) and no covariance learning (yellow) overlaid with groundtruth (red).}
	\label{fig:bicocca}
	\vspace{-6mm}
\end{figure}

We also evaluate our method on the Bicocca 25b dataset from \cite{Latif2013a}, which provides a set of odometry and loop closure constraints (represented as a pose graph) created from a bag of words place recognition system run on data collected during the Rawseeds project \cite{Ceriani2009}.
Figure \ref{fig:a} shows the loop closures in red, including many false ones.
We optimize this pose graph using our framework and show the results in Figure \ref{fig:b}. We see that our framework with the IW prior (blue) is closest to the groundtruth (red) while learning a static covariance (green) and no covariance learning (yellow) are much more negatively affected by the false loop closures. Table \ref{tab:ate} also shows the Average Trajectory Error as calculated by the Rawseeds Toolkit for each of the methods.

\begin{table}[t]
	\centering
	\vspace{5mm}
	\caption{\label{tab:ate} Average Trajectory Error as calculated by the Rawseeds Toolkit}
	\begin{tabular}{c|c|c}
		\begin{tabular}[c]{@{}c@{}@{}}IW prior (m)\end{tabular} &
		\begin{tabular}[c]{@{}c@{}@{}}Static $\mathbf{W}$ (m)\end{tabular} &
		\begin{tabular}[c]{@{}c@{}@{}}No covariance learning (m)\end{tabular}\\
		\hline
		 2.3292 & 11.9059 & 29.4418
	\end{tabular}
	 	\vspace{-4mm}
\end{table}
\vspace{-2mm}
\section{Conclusions and Future Work} \label{sec:conclusion}
In this paper, we presented parameter learning for ESGVI. We showed that our parameter learning method does not need groundtruth, and is robust to noisy measurements and outliers. This is desirable because in many cases, we do not have a way of obtaining accurate groundtruth of robot trajectories. The implication of our work is that we now have a framework for estimating robot parameters based solely on whatever sensors are available. We experimentally demonstrated our method on a 36 km vehicle dataset.

However, we still assumed two parameters to be known: $\nu$, the DOF parameter for the IW distribution, and $\beta$, the determinant constraint on the scale matrix, $\bs{\Psi}$. For future work, we will investigate how to also learn $\nu$ and eliminate the need to constrain the determinant of $\bs{\Psi}$ to be a constant, $\beta$.

In this work, we chose to learn the noise model parameters as a useful practical application of our framework. However, our future intention with ESGVI is to learn entire robot models that are represented by rich modelling techniques, such as DNNs.




%
\vspace{-2mm}
\section*{Acknowledgement}

The test vehicle used in this letter was donated by General Motors (GM) Canada.


\bibliographystyle{bib/IEEEtran}
\bibliography{bib/bib}

\end{document}